\documentclass[10pt,twocolumn,letterpaper]{article}

\usepackage{cvpr}
\usepackage{times}
\usepackage{epsfig}
\usepackage{graphicx}
\usepackage{amsmath}
\usepackage{amssymb}
\usepackage{multirow}
\usepackage{rotating}
\usepackage{color}
\usepackage{array}
\usepackage{adjustbox}
\usepackage{easytable}

\cvprfinalcopy 

\def\cvprPaperID{****} 

\begin{document}

\title{Real-world Anomaly Detection in Surveillance Videos}

\author{Waqas Sultani$^{1}$, Chen Chen$^{2}$, Mubarak Shah$^{2}$\\
$^1$Department of Computer Science, Information Technology University, Pakistan \\
$^2$Center for Research in Computer Vision (CRCV), University of Central Florida (UCF)\\
{\tt\small waqas5163@gmail.com, chenchen870713@gmail.com, shah@crcv.ucf.edu}
}

\maketitle

\begin{abstract}
Surveillance videos are able to capture a variety of realistic anomalies.
In this paper, we propose to learn anomalies by exploiting both normal and anomalous videos. To avoid annotating the anomalous segments or clips in training videos, which is very time consuming, we propose to learn anomaly through the deep multiple instance ranking framework by leveraging weakly labeled training videos, \ie the training labels (anomalous or normal) are at video-level instead of clip-level.  In our approach, we consider normal and anomalous videos as bags and video segments as instances in multiple instance learning (MIL), and automatically learn a deep anomaly ranking model that predicts high anomaly scores for anomalous video segments. Furthermore, we introduce sparsity and temporal smoothness constraints in the ranking loss function to better localize anomaly during training.

We also introduce a new large-scale first of its kind dataset of 128 hours of videos. It consists of 1900 long and untrimmed real-world surveillance videos, with 13 realistic anomalies such as fighting, road accident, burglary, robbery, etc. as well as normal activities. This dataset can be used for two tasks. First, general anomaly  detection considering all anomalies in one group and all normal activities in another group. Second, for recognizing each of 13 anomalous activities.
Our experimental results show that our MIL  method for anomaly detection achieves significant improvement on anomaly detection performance as compared to the state-of-the-art approaches.
We provide the results of several recent deep learning
baselines on anomalous activity recognition. The low recognition performance of these baselines reveals that our dataset is very challenging and opens more opportunities for future work. The dataset is available at: https://webpages.uncc.edu/cchen62/dataset.html

\end{abstract}

\section{Introduction}
Surveillance cameras are increasingly being used in public places \eg streets, intersections, banks, shopping malls, etc. to increase public safety. 
However, the monitoring capability of law enforcement agencies has not kept pace. The result is that there is a glaring deficiency in the utilization of surveillance cameras and an unworkable ratio of cameras to human monitors.
One critical task in video surveillance is detecting anomalous events such as traffic accidents, crimes or illegal activities. Generally, anomalous events rarely occur as compared to normal activities. Therefore, to alleviate the waste of labor and time, developing intelligent computer vision algorithms for automatic video anomaly detection is a pressing need. The goal of a practical anomaly detection system is to timely signal an activity that deviates normal patterns and identify the time window of the occurring anomaly. Therefore, anomaly detection can be considered as coarse level video understanding, which filters out anomalies from normal patterns. Once an anomaly is detected, it can further be categorized into one of the specific activities using classification techniques.

A small step towards addressing anomaly detection is to develop algorithms to detect a specific anomalous event, for example violence detector \cite{Mohammadi2016} and traffic accident detector \cite{kamijo2000traffic,IDM_AL}.
However, it is obvious that such solutions cannot be generalized to detect other anomalous events, therefore they render a limited use in practice.

Real-world anomalous events are complicated and diverse. It is difficult to list all of the possible anomalous events. Therefore, it is desirable that the anomaly detection algorithm does not rely on any prior information about the events. In other words, anomaly detection should be done with minimum supervision. Sparse-coding based approaches \cite{Avenue_Dataset,zhao2011online} are considered as representative methods that achieve state-of-the-art anomaly detection results. These methods assume that only a small initial portion of a video contains normal events, and therefore the initial portion is used to build the normal event dictionary. Then, the main idea for anomaly detection is that anomalous events are not accurately reconstructable from the normal event dictionary. 
However, since the environment captured by surveillance cameras can change drastically over the time (\eg at different times of a day), these approaches produce high false alarm rates for different normal behaviors.

\textbf{Motivation and contributions.} Although the above-mentioned approaches are appealing, they are based on the assumption that any pattern that deviates from the learned normal patterns would be considered as an anomaly. However, this assumption may not hold true because \textit{it is very difficult or impossible to define a normal event which takes all possible normal patterns/behaviors into account} \cite{Anomaly_Detection_Survey}. More importantly, the boundary between normal and anomalous behaviors
is often ambiguous. In addition, under realistic conditions, the same behavior could be a normal or an anomalous behavior under different conditions. Therefore, it is argued that the training data of normal and anomalous events can help an anomaly detection system learn better.
In this paper, we propose an anomaly detection algorithm using weakly labeled training videos. That is we only know the video-level labels, \ie \textit{a video is normal or contains anomaly somewhere, but we do not know where.} This is intriguing because we can easily annotate a large number of videos by only assigning video-level labels. To formulate a weakly-supervised learning approach, we resort to multiple instance learning (MIL) \cite{MIL_1997,MIL_SVM}. Specifically, we propose to learn anomaly through a deep MIL framework by treating normal and anomalous surveillance videos as bags and short segments/clips of each video as instances in a bag. Based on training videos, we automatically learn an anomaly \textit{ranking model}  that predicts high anomaly scores for anomalous segments in a video. During testing, a long-untrimmed video is divided into segments and fed into our deep network which assigns anomaly score for each video segment such that an anomaly can be detected.
In summary, this paper makes the following contributions.

$\bullet$ We propose a MIL solution to anomaly detection by leveraging only weakly labeled training videos. We propose a MIL ranking loss with sparsity and smoothness constraints for a deep learning network to learn anomaly scores for video segments. To the best of our knowledge, we are the first to formulate the video anomaly detection problem in the context of MIL. 


$\bullet$ We introduce a large-scale video anomaly detection dataset consisting of 1900 real-world surveillance videos of 13 different anomalous events and normal activities captured by surveillance cameras. It is by far the largest dataset with more than 15 times videos than existing anomaly datasets and has a total of 128 hours of videos.

$\bullet$ Experimental results on our new dataset show that our proposed method achieves superior performance as compared to the state-of-the-art anomaly detection approaches.


$\bullet$ Our dataset also serves a challenging benchmark for activity recognition on \textit{untrimmed} videos due to the complexity of activities and large intra-class variations. We provide results of baseline methods, C3D \cite{DuTran2015} and TCNN \cite{TCNN}, on recognizing 13 different anomalous activities.



\section{Related Work}

\textbf{Anomaly detection.} Anomaly detection is one of the most challenging and long standing problems in computer vision \cite{Sebe_Anomaly,ChaoticCVPR101, basharat08anomaly,Cui11abnormaldetection,Video_Parsing,Hospedales_amarkov,Context_Anomlay,USCD_Dataset,NishinoCVPR2009,Avenue_Dataset, OnlineDictioany_Anomaly,Hasan_2016_CVPR,NishinoCVPR2009}.
For video surveillance applications, there are several attempts to detect violence or aggression \cite{VioientFlow,Multi-model_voilance, Voilance_Shah,Mohammadi2016} in videos. Datta \etal proposed to detect human violence by exploiting motion and limbs orientation of people.
Kooij \etal \cite{Multi-model_voilance} employed video and audio data to detect aggressive actions in surveillance videos.  Gao \etal proposed violent flow descriptors to detect violence in crowd videos. More recently,  Mohammadi \etal \cite{Mohammadi2016} proposed a new behavior heuristic based approach to classify violent and non-violent videos.

Beyond violent and non-violent patterns discrimination, authors in \cite{ChaoticCVPR101, basharat08anomaly} proposed to use tracking to model the normal motion of people and detect deviation from that normal motion as an anomaly.  Due to difficulties in obtaining reliable tracks, several approaches avoid  tracking and learn global motion patterns through  histogram-based methods \cite{Cui11abnormaldetection}, topic modeling \cite{Hospedales_amarkov}, motion patterns \cite{saleemi-pami-2009}, social force models \cite{SocialForce},  mixtures of dynamic textures model \cite{USCD_Dataset}, Hidden Markov Model (HMM) on local spatio-temporal volumes \cite{NishinoCVPR2009}, and context-driven method \cite{Context_Anomlay}. Given the training videos of normal behaviors, these approaches learn distributions of normal motion patterns and detect low probable patterns as anomalies.

Following the success of sparse representation and dictionary learning approaches in several computer vision problems, researchers in \cite{Avenue_Dataset, OnlineDictioany_Anomaly} used sparse representation to learn the dictionary of normal behaviors.  During testing, the patterns which have large reconstruction errors are considered as anomalous behaviors.
Due to successful demonstration of deep learning for image classification, several approaches have been proposed for video action classification \cite{KarpathyCVPR14,DuTran2015}.
However, obtaining annotations for training is difficult and laborious, specifically for videos.

Recently, \cite{Hasan_2016_CVPR,Sebe_Anomaly} used deep learning based autoencoders to learn the model of normal behaviors and employed reconstruction loss to detect anomalies.
Our approach not only considers normal behaviors but also anomalous behaviors for anomaly detection, using only weakly labeled training data.

\textbf{Ranking.} Learning to rank is an active research area in machine learning. These approaches mainly focused on improving relative scores of the items instead of individual scores.  Joachims \etal \cite{RankSVM} presented rank-SVM to improve retrieval quality of search engines. Bergeron \etal \cite{MILRank} proposed an algorithm for solving multiple instance ranking problems using successive linear programming and demonstrated its application in hydrogen abstraction problem in computational chemistry. Recently, deep ranking networks have been used in several computer vision applications and have shown state-of-the-art performances.  They have been used for feature learning \cite{FineGrainedRank}, highlight detection \cite{Yao_2016_CVPR}, Graphics Interchange Format (GIF) generation \cite{Gygli_2016_CVPR}, face detection and verification \cite{FaceVerification_rank}, person re-identification \cite{PersonReIdentificantion_Rank}, place recognition \cite{Arandjelovic16}, metric learning and image retrieval \cite{Gordo2016}. All deep ranking methods require a vast amount of annotations of positive and negative samples.

In contrast to the existing methods, we formulate anomaly detection as a regression problem in the ranking framework by utilizing normal and anomalous data. To alleviate the difficulty of obtaining precise segment-level labels (\ie temporal annotations of the anomalous parts in videos) for training, we leverage multiple instance learning which relies on weakly labeled data (\ie video-level labels -- normal or abnormal, which are much easier to obtain than temporal annotations) to learn the anomaly model and detect video segment level anomaly during testing. 

%





\begin{figure*}[t]
\centering
  \includegraphics[width=0.9\linewidth]{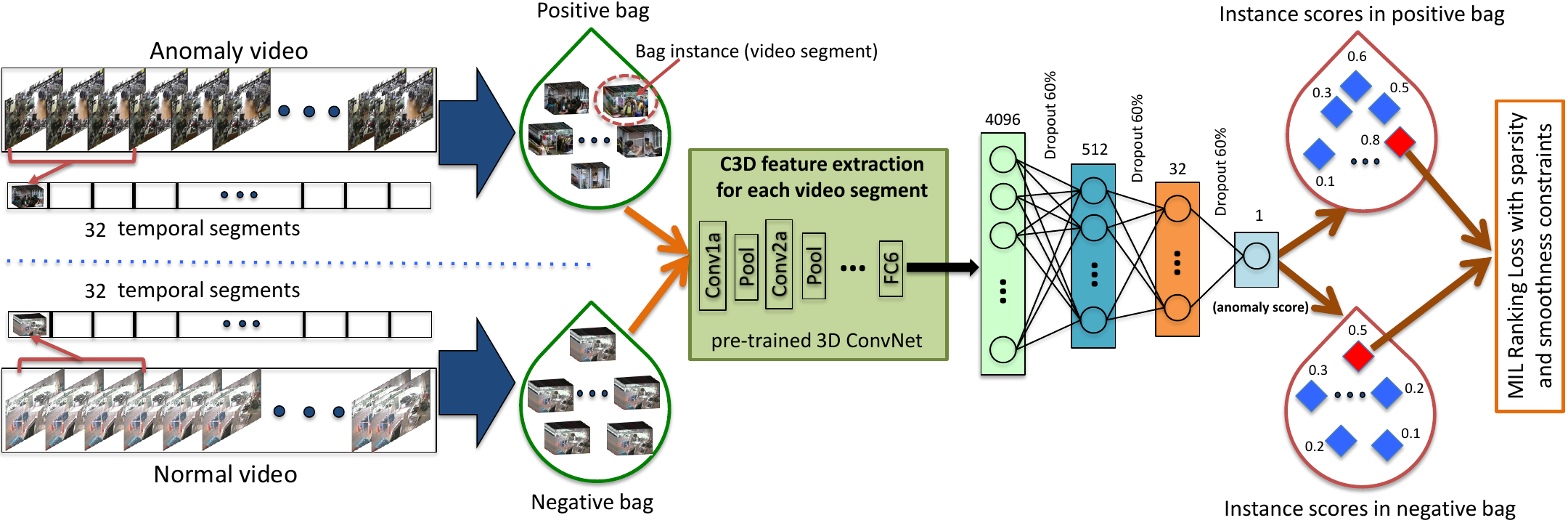}
\caption{The flow diagram of the proposed anomaly detection approach. Given the positive (containing anomaly somewhere) and negative  (containing no anomaly) videos, we divide each of them into multiple temporal video segments. Then, each video is represented as a bag and each temporal segment represents an instance in the bag. After extracting C3D features \cite{DuTran2015} for video segments, we train a fully connected neural network by utilizing a novel ranking loss function which computes the ranking loss between the highest scored instances (shown in red) in the positive bag and the negative bag.}
\label{fig:BlockDiagram1}
\end{figure*}

\section {Proposed Anomaly Detection Method}
The proposed approach (summarized in Figure \ref{fig:BlockDiagram1}) begins with dividing surveillance videos into a fixed number of segments during training. These segments make instances in a bag.  Using both positive (anomalous) and negative (normal) bags, we train the anomaly detection model using the proposed deep MIL ranking loss.

\subsection{Multiple Instance Learning}

In standard supervised classification problems using support vector machine, the labels of all positive and negative examples are available and the classifier is learned using the following optimization function:
\begin{equation}\label{eq:1}
\small
    \min_{\textbf{w} }    \phantom{x} \bigg[ \frac{1}{k}\sum_{i=1}^{k}  \overset{{\textcircled{1}}}  {\overbrace{ max(0, 1-y_i(\textbf{w}.\phi(x)-b))}} \bigg]+ \frac{1}{2}  \left \|\textbf{w} \right \|^{2},
\end{equation}
where \textcircled{1} is the hinge loss, $y_i$ represents the label of each example,  $\phi{(x)}$ denotes feature representation of an image patch or a video segment, $b$ is a bias, $k$ is the total number of training examples and $\textbf{w}$ is the classifier to be learned. To learn a robust classifier,  accurate annotations of positive and negative examples are needed. In the context of supervised anomaly detection, a classifier needs temporal annotations of each segment in videos. However, obtaining temporal annotations for videos is time consuming and laborious.

MIL relaxes the assumption of having these accurate temporal annotations. In MIL, precise temporal locations of anomalous events in videos are unknown. Instead, only video-level labels indicating the presence of an anomaly in the \textit{whole} video is needed. A video containing anomalies is labeled as positive and a video without any anomaly is labeled as negative. Then, we represent a positive video as a positive bag $\mathcal{B}_a$, where different temporal segments make individual instances in the bag, $(p^1,p^2,\ldots, p^m)$, where $m$ is the number of instances in the bag. We assume that at least one of these instances contains the anomaly. Similarly, the negative video is denoted by a negative bag, $\mathcal{B}_n$, where temporal segments in this bag form negative instances $(n^1,n^2,\ldots, n^m)$. In the negative bag, none of the instances contain an anomaly. Since the exact information (\ie instance-level label) of the positive instances is unknown, one can optimize the objective function with respect to the maximum scored instance in each bag \cite{MIL_SVM}:
\begin{equation}\label{eq:2}
\small
    \min_{\textbf{w} }   \bigg[ \frac{1}{z}\sum_{j=1}^{z} max(0, 1-Y_{\mathcal{B}_j}(\max_{i \in \mathcal{B}_j }(\textbf{w}.\phi(x_i))-b))\bigg]+\left \|\textbf{w} \right \|^{2},
\end{equation}
where $Y_{\mathcal{B}_j}$ denotes bag-level label, $z$ is the total number of bags, and all the other variables are the same as in Eq.~\ref{eq:1}.

\subsection{Deep MIL Ranking Model}


Anomalous behavior is difficult to define accurately \cite{Anomaly_Detection_Survey}, since it is quite subjective and can vary largely from person to person. Further, it is not obvious how to assign 1/0 labels to anomalies. Moreover, due to the unavailability of sufficient examples of anomaly, anomaly detection is usually treated as low likelihood pattern detection instead of classification problem \cite{Cui11abnormaldetection,Video_Parsing,Hospedales_amarkov,NishinoCVPR2009,Avenue_Dataset, OnlineDictioany_Anomaly,Hasan_2016_CVPR,NishinoCVPR2009}.

In our proposed approach, we pose anomaly detection as a regression problem.  We want the anomalous video segments to have higher anomaly scores than the normal segments.  The straightforward approach would be to use a ranking loss which encourages high scores for anomalous video segments as compared to normal segments, such as:
\begin{equation}\label{eq:3}
    f(\mathcal{V}_a)   > f(\mathcal{V}_n),
\end{equation}
where $\mathcal{V}_a$  and $\mathcal{V}_n$ represent anomalous  and normal  video segments, $f(\mathcal{V}_a)$ and $f(\mathcal{V}_n)$  represent the corresponding predicted scores, respectively.  The above ranking function should work well if the segment-level annotations are known during training.

However, in the absence of video segment level annotations, it is not possible to use Eq.~\ref{eq:3}. Instead, we propose the following multiple instance ranking objective function:
\begin{equation}\label{eq:4}
                    \max_{i \in \mathcal{B}_a} f(\mathcal{V}^i_a) >\max_{i \in \mathcal{B}_n} f(\mathcal{V}^i_n),
\end{equation}
where \textit{max} is taken over all video segments in each bag. Instead of enforcing ranking on every instance of the bag, we enforce ranking only on the two instances having the highest anomaly score respectively in the positive and negative bags. The segment corresponding to the highest anomaly score in the positive bag is most likely to be the true positive instance (anomalous segment). The segment corresponding to the highest anomaly score in the negative bag is the one looks most similar to an anomalous segment but actually is a normal instance. This negative instance is considered as a hard instance which may generate a false alarm in anomaly detection.
By using Eq.~\ref{eq:4}, we want to push the positive instances and negative instances far apart in terms of anomaly score.
Our ranking loss in the hinge-loss formulation is therefore given as follows:
\begin{equation}\label{eq:5}
  l(\mathcal{B}_a,\mathcal{B}_n)= \max(0, 1-\max_{i \in \mathcal{B}_a} f(\mathcal{V}^i_a) +\max_{i \in \mathcal{B}_n} f(\mathcal{V}^i_n)).
\end{equation}
One limitation of the above loss is that it ignores the underlying temporal structure of the anomalous video.
First, in real-world scenarios, anomaly often occurs only for a short time. In this case, the scores of the instances (segments) in the anomalous bag should be sparse, indicating only a few segments may contain the anomaly. Second, since the video is a sequence of segments, the anomaly score should vary smoothly between video segments. Therefore, we enforce temporal smoothness between anomaly scores of temporally adjacent video segments by minimizing the difference of scores for adjacent video segments.   By incorporating the sparsity and smoothness constraints on the instance scores, the loss function becomes
\begin{eqnarray}\label{eq:6}
l(\mathcal{B}_a,\mathcal{B}_n)= \max(0, 1-\max_{i \in \mathcal{B}_a} f(\mathcal{V}^i_a) +\max_{i \in \mathcal{B}_n} f(\mathcal{V}^i_n)) \nonumber\\
 +\lambda_1 \overset{{\textcircled{1}}}{\overbrace{\sum_i^{(n-1)}(f(\mathcal{V}^i_a)-f(\mathcal{V}^{i+1}_a)) ^2}} +\lambda_2 \overset{{\textcircled{2}}}{\overbrace{\sum_i^n {f(\mathcal{V}^i_a)}}},
\end{eqnarray}
where \textcircled{1} indicates the temporal smoothness term and  \textcircled{2} represents the sparsity term. In this MIL ranking loss, the error is back-propagated from the maximum scored video segments in  both  positive and negative bags. By training on a large number of positive and negative bags, we expect that the network will learn a generalized model to predict high scores for anomalous segments in positive bags (see Figure \ref{fig:Qualitative_Score_Evol_AL}). 
Finally, our complete objective function is given by
\begin{equation}\label{eq:7}
  \mathcal{L}(\mathcal{W})=  l(\mathcal{B}_a,\mathcal{B}_n) +\left \|\mathcal{W}\right \|_{F},
\end{equation}
where $\mathcal{W}$ represents model weights.

\textbf{Bags Formations.}
We divide each video into the equal number of non-overlapping temporal segments and use these video segments as bag instances.  Given each video segment, we extract the 3D convolution features \cite{DuTran2015}. We use this feature representation due to its computational efficiency, the evident capability of capturing appearance and motion dynamics in video action recognition.

\section{Dataset}

\subsection{Previous datasets}

We briefly review the existing video anomaly detection datasets. The \textbf{UMN} dataset \cite{UMN_Dataset} consists of five different staged videos where people walk around and after some time start running in different directions. The anomaly is characterized by only running action. 
\textbf{UCSD Ped1} and \textbf{Ped2} datasets \cite{USCD_Dataset} contain 70 and 28 surveillance videos, respectively. Those videos are captured at only one location. The anomalies in the videos are simple and do not reflect realistic anomalies in video surveillance, \eg people walking across a walkway, non pedestrian entities (skater, biker and wheelchair) in the walkways.
\textbf{Avenue} dataset \cite{Avenue_Dataset} consists of 37 videos. Although it contains more anomalies, they are staged and captured at one location. Similar to \cite{USCD_Dataset}, videos in this dataset are short and some of the anomalies are unrealistic (\eg throwing paper). \textbf{Subway Exit} and \textbf{Subway Entrance} datasets \cite{Subway_Dataset} contain one long surveillance video each. The two videos capture simple anomalies such as walking in the wrong direction and skipping payment. \textbf{BOSS} \cite{BOSS} dataset is collected from a surveillance camera mounted in a train. It contains anomalies such as harassment, person with a disease, panic situation, as well as normal videos. All anomalies are performed by actors.
Overall, the previous datasets for video anomaly detection are small in terms of the number of videos or the length of the video. Variations in abnormalities are also limited. In addition, some anomalies are not realistic.

\subsection{Our dataset}

Due to the limitations of previous datasets, we construct a new large-scale dataset to evaluate our method. It consists of long \textit{untrimmed surveillance videos} which cover 13 real-world anomalies, including \textit{Abuse, Arrest, Arson, Assault, Accident, Burglary, Explosion,  Fighting, Robbery, Shooting,  Stealing, Shoplifting,} and \textit{Vandalism}. These anomalies are selected because they have a significant impact on public safety. We compare our dataset with previous anomaly detection datasets in Table \ref{tab:dataset_comparison}.

\begin{table*}
\begin{center}
\footnotesize
\begin{tabular}{|l|l|l|l|l|}
\hline
                & \# of videos & Average frames & Dataset length & Example anomalies                                                                                             \\ \hline
UCSD Ped1 \cite{USCD_Dataset}      & \phantom{333}70           & \phantom{333}201             & \phantom{333}5 min          & \begin{tabular}[c]{@{}l@{}}Bikers, small carts, walking across walkways\end{tabular}                 \\ \hline
UCSD Ped2  \cite{USCD_Dataset}     & \phantom{333}28           & \phantom{333}163             & \phantom{333}5 min          & \begin{tabular}[c]{@{}l@{}}Bikers, small carts, walking across walkways\end{tabular}                \\ \hline
Subway Entrance  \cite{Subway_Dataset}  & \phantom{333}1            & \phantom{333}121,749         & \phantom{333}1.5 hours      & Wrong direction, No payment                                                                           \\ \hline
Subwa Exit \cite{Subway_Dataset}     & \phantom{333}1            & \phantom{333}64,901          & \phantom{333}1.5 hours      & Wrong direction, No payment                                                                           \\ \hline
Avenue \cite{Avenue_Dataset}         & \phantom{333}37           &\phantom{333} 839             & \phantom{333}30 min         & Run, throw, new object                                                                                \\ \hline
UMN \cite{UMN_Dataset}             & \phantom{333}5            & \phantom{333}1290            & \phantom{333}5 min          & Run                                                                                                   \\ \hline
BOSS  \cite{BOSS}           & \phantom{333}12            & \phantom{333}4052            & \phantom{333}27 min          & Harass, Disease, Panic                                          \\ \hline
\textbf{Ours}            & \phantom{33}\bf{1900}         & \phantom{333}\bf{7247}            & \phantom{333}\bf{128 hours}      & \begin{tabular}[c]{@{}l@{}}\textbf{Abuse, arrest, arson, assault, accident, burglary, fighting, robbery} \end{tabular} \\ \hline
\end{tabular}
\vspace{1em}
\caption{A comparison of anomaly datasets. Our dataset contains larger number of longer surveillance videos with more realistic anomalies. }
\label{tab:dataset_comparison}
\end{center}
\end{table*}

\begin{figure*}
\centering
  \includegraphics[width=0.98\linewidth]{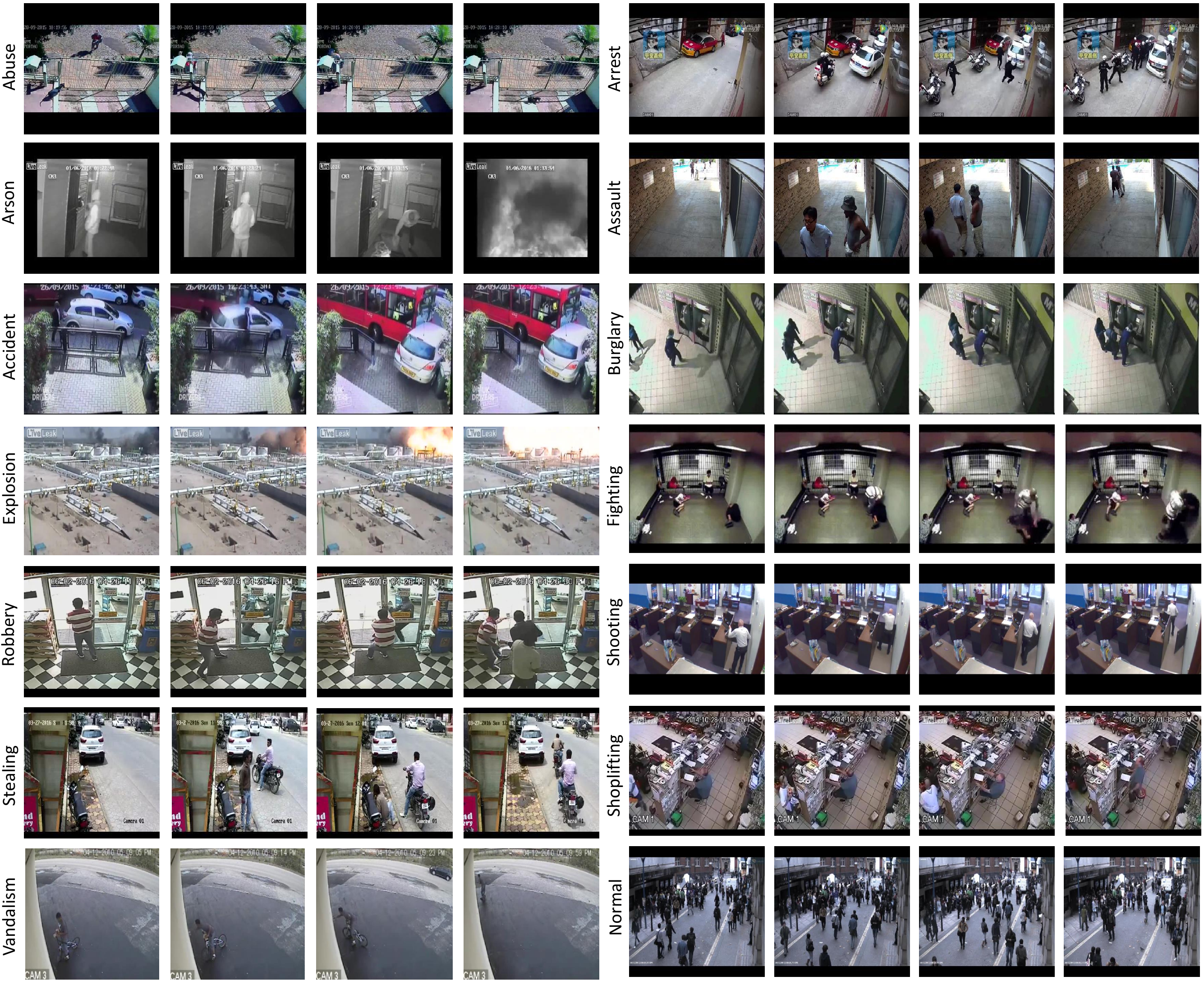}
\caption{Examples of different anomalies from the training and testing videos in our dataset.}
\label{fig:Dataset_pic}
\end{figure*}

\textbf{Video collection.} To ensure the quality of our dataset, we train ten annotators (having different levels of computer vision expertise) to collect the dataset.
We search videos on YouTube\footnote{https://www.youtube.com/} and LiveLeak\footnote{https://www.liveleak.com/} using text search queries (with slight variations \eg ``car crash", ``road accident") of each anomaly. In order to retrieve as many videos as possible, we also use text queries in different languages (\eg French, Russian, Chinese, etc.) for each anomaly, thanks to Google translator.
We remove videos which fall into any of the following conditions: manually edited, prank videos, not captured by CCTV cameras, taking from news, captured using a hand-held camera, and containing compilation.  We also discard videos in which the anomaly is not clear. With the above video pruning constraints, 950 unedited real-world surveillance videos with clear anomalies are collected. Using the same constraints, 950 normal videos are gathered, leading to a total of 1900 videos in our dataset. In Figure~\ref{fig:Dataset_pic}, we show four frames of an example video from each anomaly.

\textbf{Annotation.} 
For our anomaly detection method, only video-level labels are required for training. However, in order to evaluate its performance on testing videos, we need to know the temporal annotations, \ie the start and ending frames of the anomalous event in each testing anomalous video.
To this end, we assign the same videos to multiple annotators to label the temporal extent of each anomaly. The final temporal annotations are obtained by averaging annotations of different annotators. The complete dataset is finalized after intense efforts of several months.

\begin{figure}
\centering
  \includegraphics[width=8.3cm,height=3.2cm]{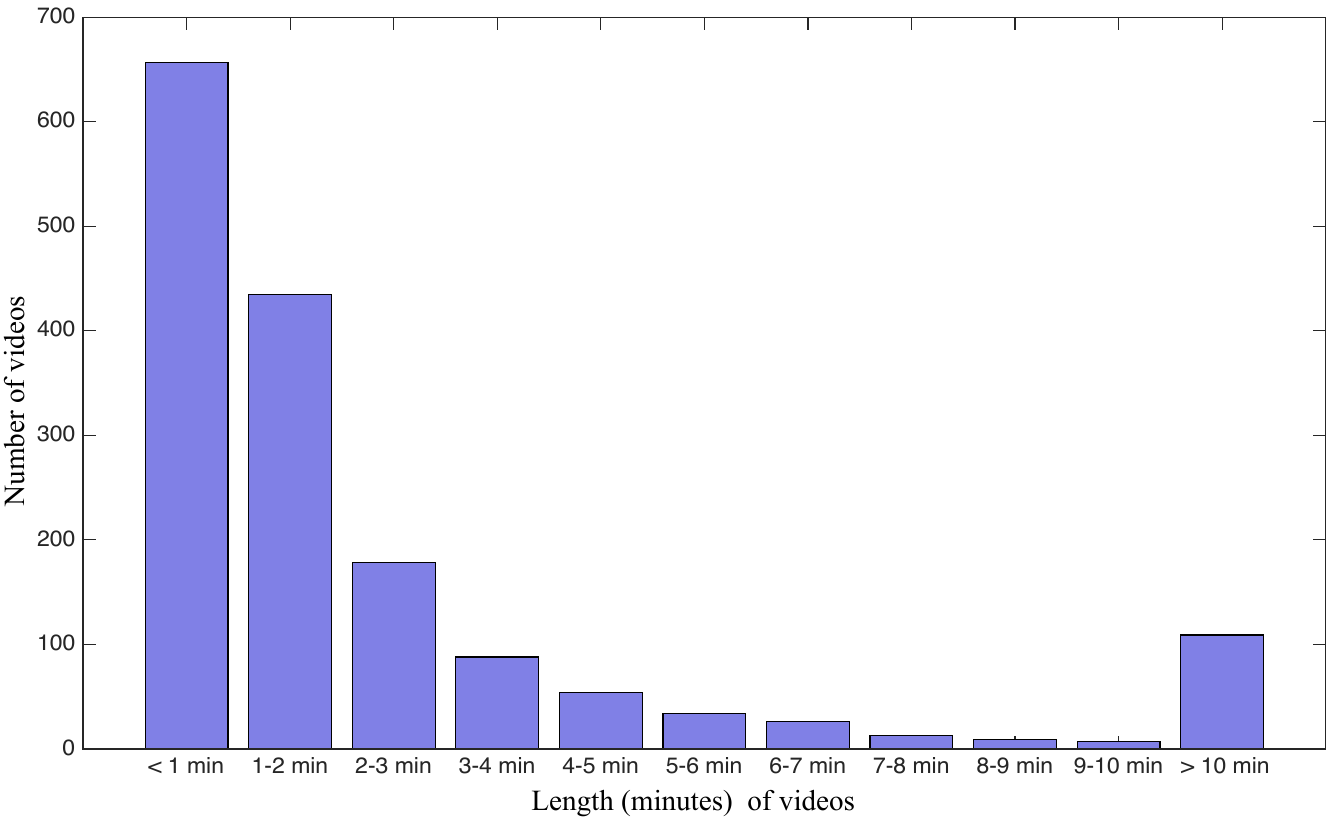}
\caption{Distribution of videos according to length (minutes) in the training set.}
\label{fig:length_Distribuition}
\end{figure}

\begin{figure}
\centering
  \includegraphics[width=8.3cm,height=3.2cm]{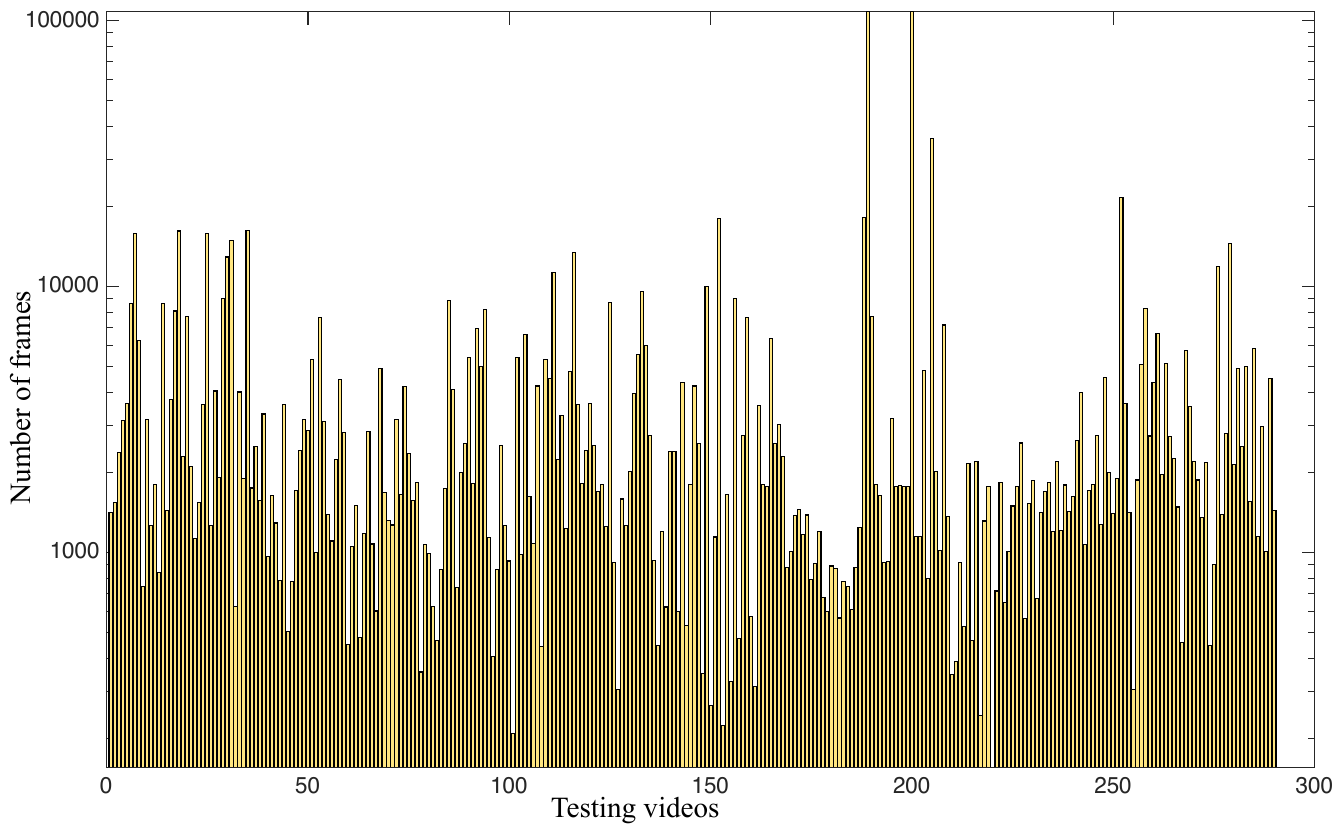}
\caption{Distribution of video frames in the testing set.}
\label{fig:frames_Distribuition}
\end{figure}

\begin{figure}
\centering
  \includegraphics[width=8.3cm,height=3.2cm]{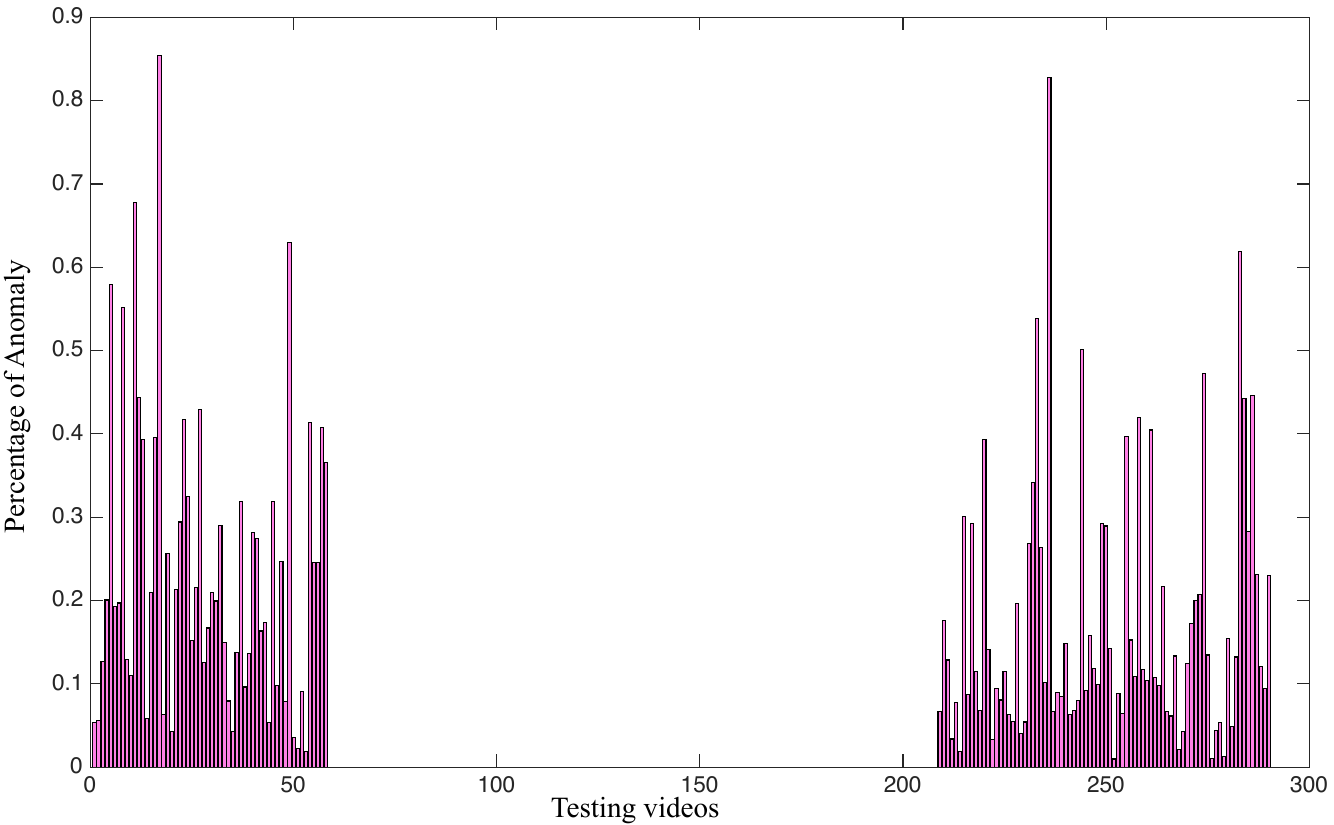}
\caption{Percentage of anomaly in each video of the testing set. Normal videos (59 to 208) do not contain any anomaly.}
\label{fig:Anomaly_Distribuition}
\end{figure}

\textbf{Training and testing sets.} We divide our dataset into two parts: the training set consisting of 800 normal and 810 anomalous videos (details shown in Table \ref{tab:video_number}) and the testing set including the remaining 150 normal and 140 anomalous videos. Both training and testing sets contain all 13 anomalies at various temporal locations in the videos. Furthermore, some of the videos have multiple anomalies. The distribution of the training videos in terms of length (in minute) is shown in Figures \ref{fig:length_Distribuition}.
The number of frames and percentage of anomaly in each testing video are presented in Figures \ref{fig:frames_Distribuition} and~\ref{fig:Anomaly_Distribuition}, respectively.



\begin{table}
\begin{center}
\footnotesize
\begin{tabular}{p{0.1cm}|p{0.1cm}p{0.1cm}p{0.1cm}p{0.1cm}p{0.1cm}p{0.1cm}p{0.1cm}p{0.1cm}p{0.1cm}p{0.1cm}p{0.1cm}p{0.1cm}p{0.1cm}||p{0.1cm}}

\rotatebox{90}{\textbf{Anomaly}}      & \rotatebox{90}{Abuse} &  \rotatebox{90}{Arrest} & \rotatebox{90}{Arson} &  \rotatebox{90}{Assault} &  \rotatebox{90}{Burglary} &  \rotatebox{90}{Explosion} &   \rotatebox{90}{Fighting} &  \rotatebox{90}{Road Accidents} & \rotatebox{90}{Robbery} & \rotatebox{90}{Shooting} & \rotatebox{90}{Shoplifting} & \rotatebox{90}{Stealing} & \rotatebox{90}{Vandalism} & \rotatebox{90}{\textbf{Normal events}} \\
\hline
\rotatebox{90}{\textbf{\# of videos}} &
\rotatebox{90}{\vspace{2em} 50  {(48)}  }   & \rotatebox{90}{\vspace{2em} 50  {(45)}   }     & \rotatebox{90}{\vspace{2em} 50  {(41)}  }    & \rotatebox{90}{\vspace{2em} 50  {(47)}  }  & \rotatebox{90}{\vspace{2em} 100  {(87)}  }    & \rotatebox{90}{\vspace{2em} 50   {(29)}  }     & \rotatebox{90}{\vspace{2em} 50   {(45)}  }    & \rotatebox{90}{\vspace{2em} 150  {(127)}  } &  \rotatebox{90}{\vspace{2em} 150   {(145)} }    & \rotatebox{90}{\vspace{2em} 50   {(27)}}     & \rotatebox{90}{\vspace{2em} 50     {(29)}}    & \rotatebox{90}{\vspace{2em} 100   {(95)}} & \rotatebox{90}{\vspace{2em} 50   {(45)}} & \rotatebox{90}{\vspace{2em} 950   {(800)}}\\

\end{tabular}
\end{center}
\caption{Number of videos of each anomaly in our dataset. Numbers in  brackets represent the number of videos in the training set.}
\label{tab:video_number}
\end{table}


\section{Experiments}
\subsection{Implementation Details}

We extract visual features from the fully connected (FC) layer FC6 of the C3D network \cite{DuTran2015}.  Before computing features, we re-size each video frame to 240 $\times$ 320 pixels and fix the frame rate to 30 fps. We compute C3D features for every 16-frame video clip followed by $l_2$ normalization. To obtain features for a video segment, we take the average of all 16-frame clip features within that segment.  We input these features (4096D) to a  3-layer FC neural network. The first FC layer has 512 units followed by 32 units and 1 unit FC layers. 60\% dropout regularization \cite{Dropout} is used between FC layers. We experiment with deeper networks but do not observe better detection accuracy.
We use ReLU \cite{Relu} activation and Sigmoid activation for the first and the last FC layers respectively, and employ Adagrad \cite{Adagrad} optimizer with the initial learning rate of 0.001.
The parameters of sparsity and smoothness constraints in the MIL ranking loss are set to  $\lambda_1$=$\lambda_2$ = $8 \times 10^{-5}$ for the best performance.

We divide each video into 32 non-overlapping segments and consider each video segment as an instance of the bag. The number of segments (32) is empirically set. We also experimented with multi-scale overlapping temporal segments but it does not affect detection accuracy. We randomly select 30 positive and 30 negative bags as a mini-batch. We compute gradients by reverse mode automatic differentiation on computation graph using Theano \cite{THEANO}. Specifically, we identify set of variables on which loss depends, compute gradient for each variable and obtain final gradient through chain rule on the computation graph. Each video passes through the network and we get the score for each of its temporal segments. Then we compute loss as shown in Eq.~\ref{eq:6} and Eq.~\ref{eq:7} and back-propagate the loss for the whole batch.

\textbf{Evaluation Metric.}
Following previous works on anomaly detection \cite{USCD_Dataset}, we use frame based receiver operating characteristic (ROC) curve and corresponding area under the curve (AUC) to evaluate the performance of our method. We do not use equal error rate (EER) \cite{USCD_Dataset} as it does not measure anomaly correctly, specifically if only a small portion of a long video contains anomalous behavior.




\subsection{Comparison with the State-of-the-art}



We compare our method with two state-of-the-art approaches for anomaly detection. Lu \etal \cite{Avenue_Dataset} proposed \textbf{dictionary based approach} to learn the normal behaviors and used reconstruction errors to detect anomalies. Following their code, we extract 7000 cuboids from each of the normal training video and compute gradient based features in each volume. After reducing the feature dimension using PCA, we learn the dictionary using sparse representation. Hasan \etal \cite{Hasan_2016_CVPR} proposed a fully convolutional feed-forward \textbf{deep auto-encoder based approach} to learn local features and classifier. Using their implementation, we train the network on normal videos using the temporal window of 40 frames.  Similar to \cite{Avenue_Dataset}, reconstruction error is used to measure anomaly.
We also use a \textbf{binary SVM classifier} as a baseline method. Specifically, we treat all anomalous videos as one class and normal videos as another class. C3D features are computed for each video, and a binary classifier is trained with linear kernel. For testing, this classifier provides the probability of each video clip to be anomalous. 



\begin{figure}
\centering
  \includegraphics[width=6.0cm,height=5cm]{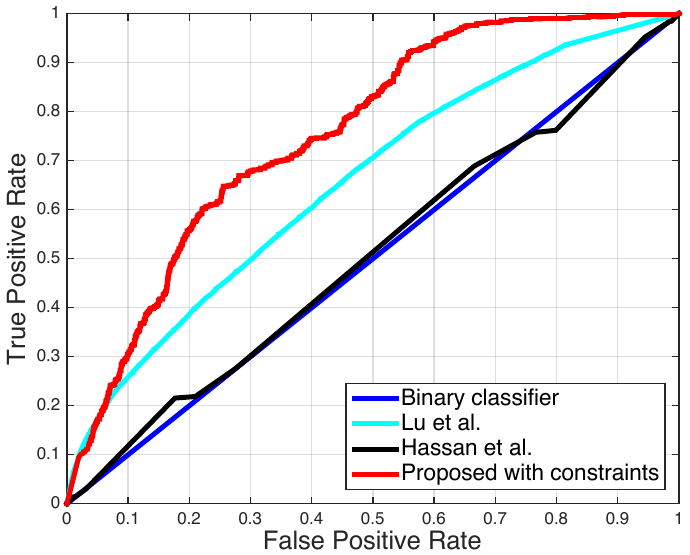}
\caption{ROC comparison of binary classifier (blue), Lu \etal \cite{Avenue_Dataset} (cyan), Hasan \etal \cite{Hasan_2016_CVPR} (black), proposed method without constraints (magenta) and with constraints (red).}
\label{fig:ROC_Comparision_AL}
\end{figure}

\begin{figure*}
\centering
  \includegraphics[width=0.99\linewidth]{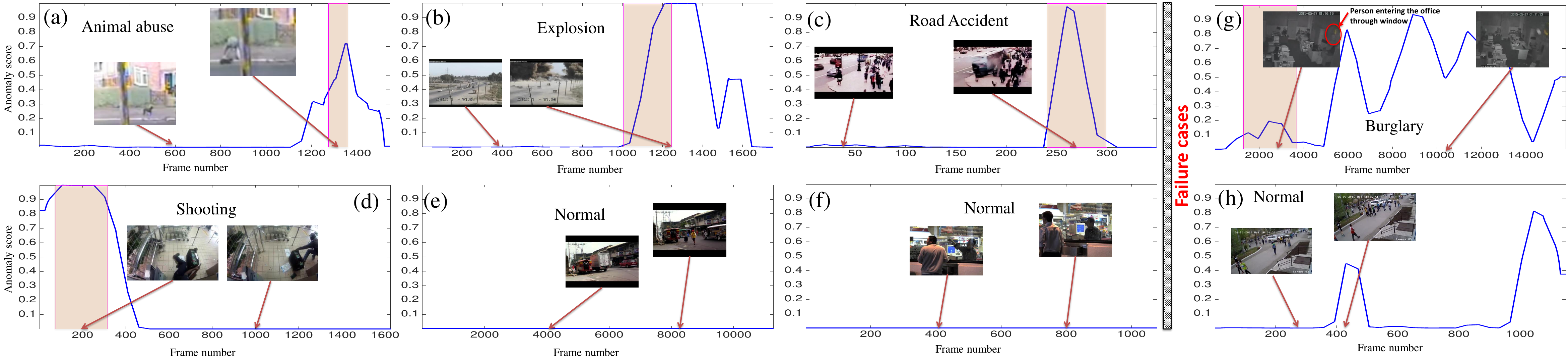}
\caption{Qualitative results of our method on testing videos. Colored window shows ground truth anomalous region. (a), (b), (c) and (d) show videos containing \textit{animal abuse (beating a dog)}, \textit{explosion}, \textit{road accident} and \textit{shooting}, respectively. (e) and (f) show normal videos with no anomaly. (g) and (h) present two failure cases of our anomaly detection method.
}
\label{fig:Qualitative_Testing_Score3}
\end{figure*}

The quantitative comparisons in terms of ROC and AUC are shown in  Figure \ref{fig:ROC_Comparision_AL} and Table \ref{table:Comparision_AL}. We also compare the results of our approach with and without smoothness and sparsity constraints. The results show that our approach significantly outperforms the existing methods. Particularly, our method achieves much higher true positive rates than other methods under low false positive rates \eg 0.1-0.3.



The binary classifier results demonstrate that traditional action recognition approaches cannot be used for anomaly detection in real-world surveillance videos.  This is because our dataset contains long untrimmed videos where anomaly mostly occurs for a short period of time.  Therefore, the features extracted from these untrimmed training videos are not discriminative enough for the anomalous events. In the experiments, binary classifier produces very low anomaly scores for almost all testing videos. 
Dictionary learnt by \cite{Avenue_Dataset} is not robust enough to discriminate between normal and anomalous pattern. In addition to producing the low reconstruction error for normal portion of the videos, it also produces low reconstruction error for anomalous part.
Hasan \etal \cite{Hasan_2016_CVPR} learns normal patterns quite well. However, it tends to produce high anomaly scores even for new normal patterns.
Our method performing significantly better than \cite{Hasan_2016_CVPR} demonstrates the effectiveness and it emphasizes that training using both anomalous and normal videos are indispensable for a robust anomaly detection system.

In Figure \ref{fig:Qualitative_Testing_Score3}, we present qualitative results of our approach on eight videos. (a)-(d) show four videos with anomalous events. Our method provides successful and timely detection of those anomalies by generating high anomaly scores for the anomalous frames. (e) and (f) are two normal videos. Our method produces low anomaly scores (close to 0) through out the entire video, yielding zero false alarm for the two normal videos. We also illustrate two failure cases in (g) and (h). Specifically, (g) is an anomalous video containing a burglary event (person entering an office through a window). Our method fails to detect the anomalous part because of the darkness of the scene (a night video). Also, it generates false alarms mainly due to occlusions by flying insects in front of camera. In (h), our method produces false alarms due to sudden people gathering (watching a relay race in street). In other words, it fails to identify the normal group activity.



\tabcolsep=1cm
\begin{table}
\begin{center}
\small{\begin{tabular}{|l|c|l}
\hline
Method & AUC    \\
\hline\hline
Binary classifier  & 50.0  \\
\hline
Hasan \etal \cite{Hasan_2016_CVPR} & 50.6  \\
\hline
Lu \etal \cite{Avenue_Dataset} & 65.51 \\
\hline
Proposed w/o constraints  & 74.44 \\
\hline
\textbf{Proposed w constraints}   & \textbf{75.41}\\
\hline
\end{tabular}}
\end{center}
\caption{AUC comparison of various approaches on our dataset.}\label{table:Comparision_AL}
\end{table}

\subsection{Analysis of the Proposed Method}

\begin{figure}[t]
\centering
  \includegraphics[width=0.98\columnwidth]{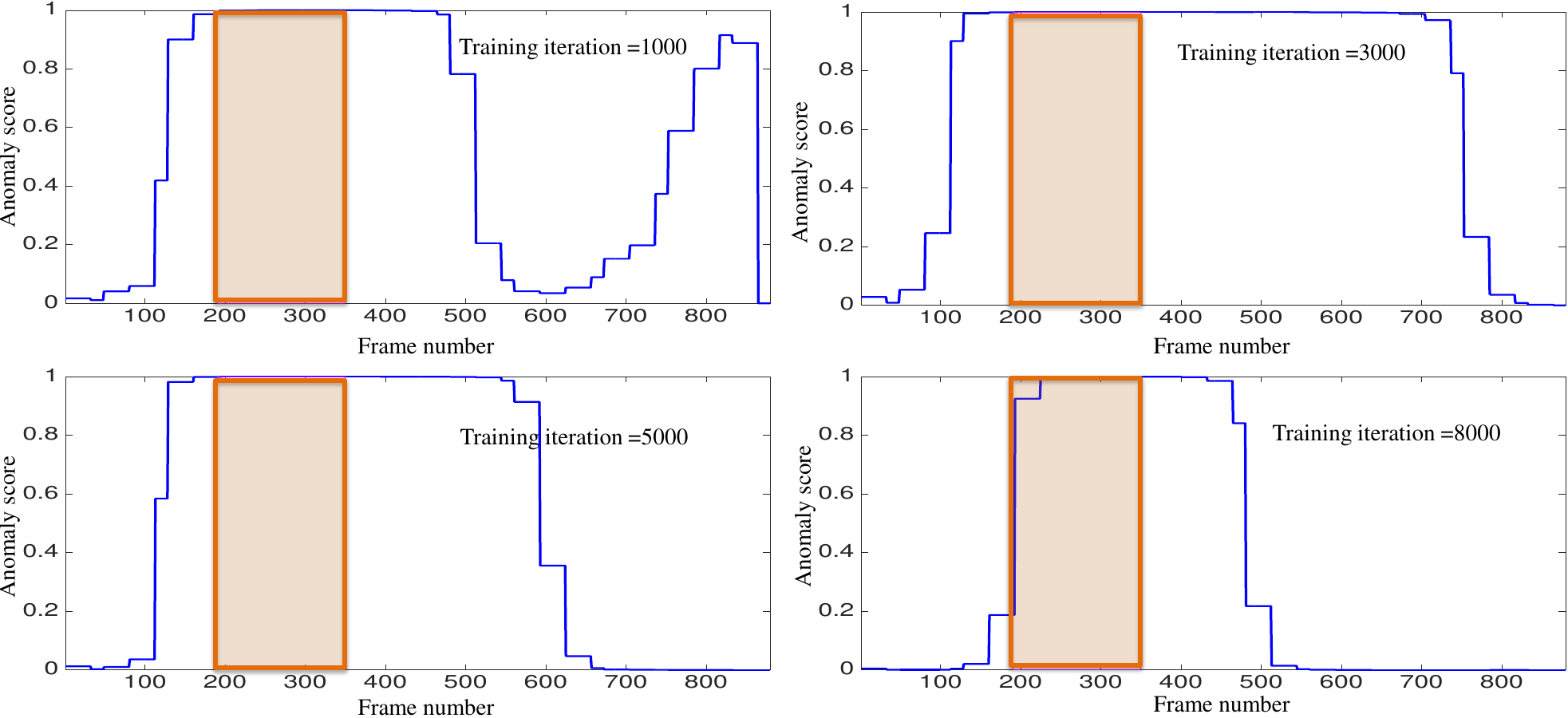}
\caption{Evolution of score on a training video over iterations. Colored window represents ground truth (anomalous region).  As iteration increases, our method generates high anomaly scores on anomalous video segments and low scores on normal segments.}
\label{fig:Qualitative_Score_Evol_AL}
\end{figure}

\textbf{Model training.} The underlying assumption of the proposed approach is that given a lot of positive and negative videos with video-level labels, the network can automatically learn to predict the location of the anomaly in the video. To achieve this goal, the network should learn to produce high scores for anomalous video segments during training iterations. Figure  \ref{fig:Qualitative_Score_Evol_AL} shows the evolution of anomaly score for a training anomalous example over the iterations. At 1,000 iterations, the network predicts high scores for both anomalous and normal video segments. After 3,000 iterations, the network starts to produce low scores for normal segments and keep high scores of anomalous segments. As the number of iterations increases and the network sees more videos, it automatically learns to precisely localize anomaly.  Note that although we do not use any segment level annotations, the network is able to predict the temporal location of an anomaly in terms of anomaly scores. 


\textbf{False alarm rate.} In real-world setting, a major part of a surveillance video is normal. A robust anomaly detection method should have low false alarm rates on normal videos. Therefore, we evaluate the performance of our approach and other methods on normal videos only. Table \ref{FalseAlarm} lists the false alarm rates of different approaches at 50\% threshold. Our approach has a much lower false alarm rate than other methods, indicating a more robust anomaly detection system in practice. This validates that using both anomalous and normal videos for training helps our deep MIL ranking model to learn more general normal patterns.

\tabcolsep=0.5cm
\begin{table}
\begin{center}
\footnotesize{
\centering
\begin{tabular}{|c|c|c|c|c|}
\hline
Method & \cite{Hasan_2016_CVPR} &  \cite{Avenue_Dataset}  & \textbf{Proposed} \\ \hline \hline
False alarm rate & 27.2         & 3.1              &  \textbf{1.9}      \\ \hline
\end{tabular}}
\end{center}
\caption{False alarm rate comparison on normal testing videos.} 
\label{FalseAlarm}
\end{table}

\subsection{Anomalous Activity Recognition Experiments}
\begin{figure}
\centering
  \includegraphics[width=8.3cm,height=3.8cm]{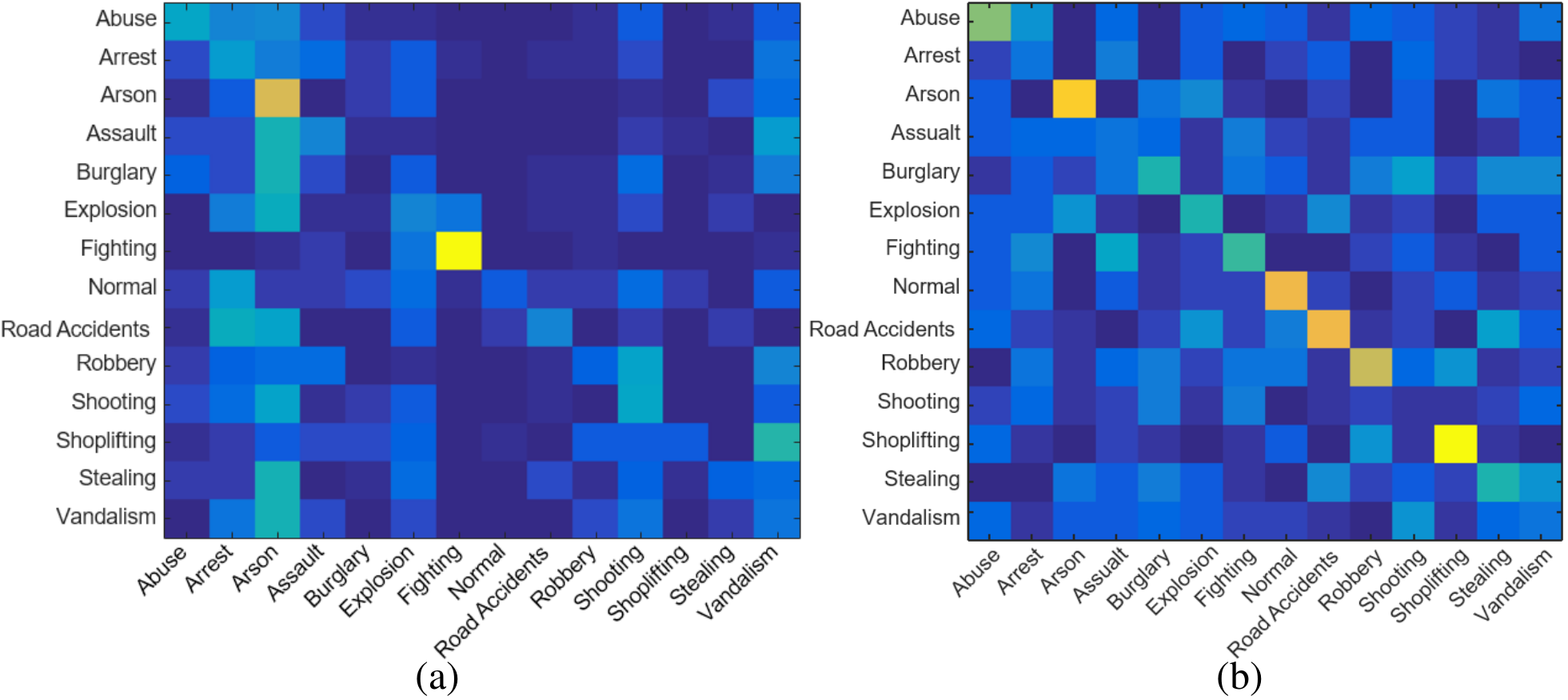}
\caption{(a) and (b) show  the confusion matrices of activity recognition using C3D \cite{DuTran2015} and TCNN \cite{TCNN} on our dataset.} 
\label{fig:action_matrices}
\end{figure}

Our dataset can be used as an anomalous activity recognition benchmark since we have event labels for the anomalous videos during data collection, but which are not used for our anomaly detection method discussed above. For activity recognition, we use 50 videos from each event and divide them into 75/25 ratio for training and testing\footnote{Training/testing partitions will be made publicly available.}. We provide two baseline results for activity recognition on our dataset based on a 4-fold cross validation. For the first baseline, we construct a 4096-D feature vector by averaging C3D \cite{DuTran2015} features from each 16-frames clip followed
by an L2-normalization. The feature vector is used as input to a nearest neighbor classifier. The second baseline is the Tube Convolutional Neural Network (TCNN) \cite{TCNN}, which introduces the tube of interest (ToI) pooling layer to replace the 5-th 3d-max-pooling layer in C3D pipeline. The ToI pooling layer aggregates features from all clips and outputs one feature vector for a whole video. Therefore, it is an end-to-end deep learning based video recognition approach.
The quantitative results \ie confusion matrices and accuracy are given in Figure \ref{fig:action_matrices} and Table \ref{tab:Action_Accuracy}. These state-of-the-art action recognition methods perform poor on this dataset. It is because the videos are long untrimmed surveillance videos with low resolution. In addition, there are large intra-class variations due to changes in camera viewpoint and illumination, and background noise. Therefore, our dataset is a unique and challenging dataset for anomalous activity recognition.

\tabcolsep=0.6cm
\begin{table}[h]
\begin{center}
\footnotesize{
\centering
\begin{tabular}{|c|c|c|}
\hline
Method & C3D \cite{DuTran2015} &  TCNN \cite{TCNN} \\ \hline
Accuracy & 23.0 &   28.4 \\ \hline
\end{tabular}}
\end{center}
\caption{Activity recognition results of C3D \cite{DuTran2015} and TCNN \cite{TCNN}.}
\label{tab:Action_Accuracy}
\end{table}

\section{Conclusions}
We propose a deep learning approach to detect real-world anomalies in surveillance videos. Due to the complexity of these realistic anomalies, using only normal data alone may not be optimal for anomaly detection. We attempt to exploit both normal and anomalous surveillance videos. To avoid labor-intensive temporal annotations of anomalous segments in training videos, we learn a general model of anomaly detection using deep multiple instance ranking framework with weakly labeled data. To validate the proposed approach, a new large-scale anomaly dataset consisting of a variety of real-world anomalies is introduced. The experimental results on this dataset show that our proposed anomaly detection approach performs significantly better than baseline methods. Furthermore, we demonstrate the usefulness of our dataset for the second task of anomalous activity recognition.

\section{Acknowledgement}
The project was supported by Award No.
2015-R2-CX-K025, awarded by the National Institute of Justice,
Office of Justice Programs, U.S. Department of Justice. The opinions,
findings, and conclusions or recommendations expressed in
this publication are those of the author(s) and do not necessarily
reflect those of the Department of Justice.

{\small
\bibliographystyle{ieee}
\bibliography{egbib}
}

\end{document}